\documentclass[conference,a4paper]{IEEEtran}
\IEEEoverridecommandlockouts

\usepackage[hidelinks]{hyperref}
\usepackage[cmex10]{amsmath}%American Math Society(AMS) math formatting
\usepackage{amssymb,amsfonts}%AMS extra symbols and fonts
\usepackage{dblfloatfix}%fix double column figure ordering and placement

\usepackage[ruled,vlined]{algorithm2e}
\usepackage{graphicx}
\graphicspath{{Figures/PDF/}{Figures/PNG/}}

\usepackage{booktabs}
\usepackage{siunitx}
\usepackage[numbers,compress]{natbib}
\usepackage{texnames}
\usepackage{bm,bbm}
\usepackage{orcidlink}
\usepackage{multirow}

\begin{document}

\title{\uppercase{PF-Trans: Physics-embedded Frequency-aware Transformer for Spectral Reconstruction}
\thanks{This work was supported by the National Natural Science Foundation of China for the Major Scientific Instrument Development Project under Grant No. 62327803}
}

\author{	\IEEEauthorblockN{Yuzhe Gui\orcidlink{0009-0007-2630-6736}}
	\IEEEauthorblockA{\textit{Harbin Institute of Technology}\\
		150001 Harbin, China\\
		25B305002@stu.hit.edu.cn}
    \and
	\IEEEauthorblockN{Tianzhu Liu}
	\IEEEauthorblockA{\textit{Harbin Institute of Technology}\\
		150001 Harbin, China\\
		tzliu@hit.edu.cn}
    \and
	\IEEEauthorblockN{Yanfeng Gu}
	\IEEEauthorblockA{\textit{Harbin Institute of Technology}\\
		150001 Harbin, China\\
		guyf@hit.edu.cn}
    \and
	\IEEEauthorblockN{Xian Li}
	\IEEEauthorblockA{\textit{Harbin Institute of Technology}\\
		150001 Harbin, China\\
		xianli@hit.edu.cn}
}

\maketitle
\begin{abstract}
Snapshot Broadband Filter Array (BFA) imaging provides high light throughput for spectral reconstruction but introduces severe spectral aliasing due to complex modulation. Current deep learning approaches, limited to spatial denoising, often fail to address the global frequency-specific degradations caused by the mask structure. To address this, we propose a Physics-embedded Frequency-aware Transformer (PF-Trans) for high-fidelity remote sensing spectral reconstruction. Our method explicitly integrates the physical sensing model through mask injection and a gray-scale consistency loss to ensure physical fidelity. Furthermore, we introduce a Dual-domain Block with a parallel Fast Fourier Transform (FFT) branch, enabling the network to perceive and suppress aliasing artifacts in the frequency domain. Extensive experiments on multiple datasets demonstrate that PF-Trans achieves state-of-the-art performance, achieving a Peak Signal-to-Noise Ratio (PSNR) of up to 48.50 dB on the GF-5 Shanghai dataset, significantly outperforming comparison methods.
\end{abstract}

\begin{IEEEkeywords}
Remote Sensing Images; Computational Imaging; Broadband Filter Array; Physics-driven Deep Learning; Hyperspectral Reconstruction
\end{IEEEkeywords}

\section{Introduction}

Hyperspectral imaging (HSI) captures continuous spectral signatures, enabling precise material identification, which is critical in remote sensing and biomedical fields \cite{Li2019, li2020deep, li2023end, qin2024hemisphere}.
However, conventional push-broom scanning systems rely on strictly synchronized line-by-line scanning, which suffers from low temporal resolution and susceptibility to platform vibrations. This limits their ability to capture dynamic scenes or be deployed on lightweight platforms such as UAVs. To address this, Snapshot Compressive Imaging (SCI), such as Coded Aperture Snapshot Spectral Imaging (CASSI) \cite{Wagadarikar2008}, has been developed to capture 3D spectral cubes in a single shot. Among SCI techniques, the Broadband Filter Array (BFA) proposed by Bian et al. \cite{Bian2024} stands out for its high light throughput and dynamic range compared to its narrowband counterparts, making it a promising candidate for next-generation optical remote sensing payloads where the signal-to-noise ratio (SNR) is paramount.

Recovering the HSI cube from the compressed BFA measurement is a challenging ill-posed problem. Although compressive sensing theory \cite{Duarte2011} provides mathematical limits, traditional model-based reconstruction methods rely on iterative optimization, which is often computationally prohibitive for large-scale remote sensing data. Recently, Deep Learning (DL) approaches \cite{Shi2018, Stiebel2018, Xiong2017, li2022spectral, li2024multi} 
have achieved promising results by learning end-to-end mappings. State-of-the-art methods, such as MST++ \cite{Cai2022}, employ Transformers to model global dependencies. However, these methods typically operate exclusively in the spatial domain and treat reconstruction as blind denoising \cite{Zhang2020, Hu2021}.
They fail to effectively decouple the complex aliasing patterns introduced by the periodic BFA mask from the high-frequency texture details of ground objects. Consequently, this leads to residual grid-like artifacts and spectral distortions, which severely hamper the interpretation of fine-grained ground features in complex urban scenes.

To overcome these limitations and tailor BFA reconstruction for remote sensing, we propose a Physics-embedded Frequency-aware Transformer (PF-Trans). By synergizing physical constraints with frequency-domain processing, our framework addresses the specific challenge of mask-induced spectral aliasing. Our main contributions are summarized as follows:
% [Figure 1 Placement]
\begin{figure*}[t!]
	\centering
	\includegraphics[width=0.95\textwidth]{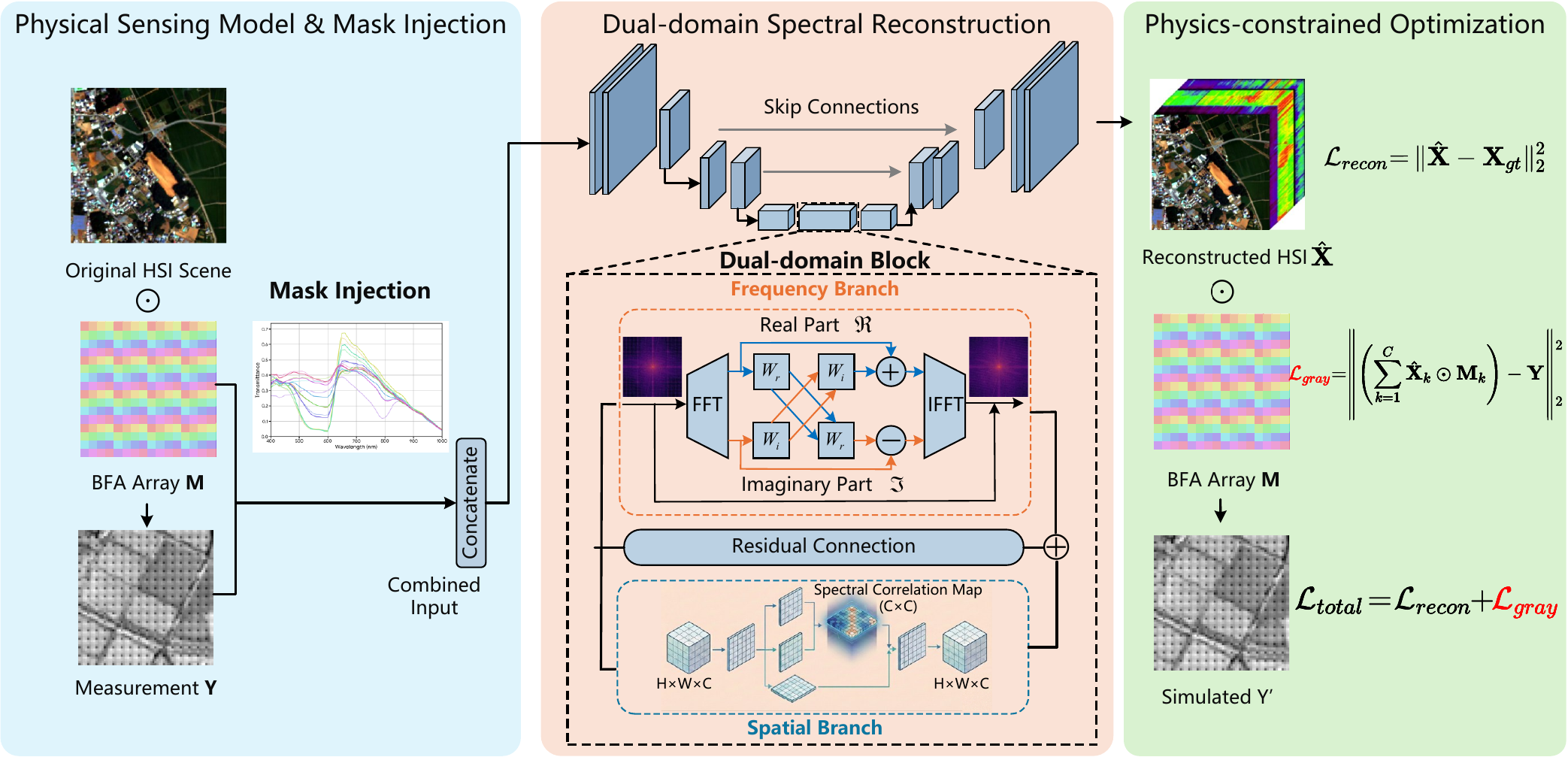}
	\caption{Overview of the proposed framework. Integrating mask-embedded input, dual-domain reconstruction, and closed-loop consistency loss, our physics-driven architecture ensures high-fidelity and optically plausible hyperspectral recovery.}
	\label{fig:architecture}
\end{figure*}
\begin{itemize}
    \item We devise a Dual-domain Block incorporating a parallel FFT branch. By exploiting the global receptive field, this mechanism effectively decouples periodic mask aliasing from ground textures, transforming artifact removal into a precise spectral filtering process.
    \item We explicitly integrate the physical sensing model via mask injection and a gray-scale consistency loss to ensure physical fidelity, bridging the gap between data-driven priors and the actual BFA optical path.
    \item We conduct comprehensive evaluations across diverse scenarios, spanning satellite, airborne, and UAV platforms. The results demonstrate that PF-Trans not only establishes new state-of-the-art benchmarks but also exhibits superior generalization capabilities in complex real-world applications.
\end{itemize}

\section{Methodology}

We propose a physics-driven framework that unifies optical sensing with deep learning to address the ill-posed nature of compressive imaging. As shown in Fig. \ref{fig:architecture}, our approach transforms blind estimation into informed decoding by explicitly injecting the physical mask as an inductive bias. To handle complex aliasing, we employ a dual-domain backbone synergizing spatial attention with frequency filtering. Furthermore, a closed-loop consistency loss is enforced, ensuring the reconstructed hyperspectral cube remains optically consistent with the raw measurements.

\subsection{Physical Sensing Model and Mask Injection}
\label{sec:sensing_model}

\subsubsection*{1) Continuous Optical Process}
In a practical deployment scenario, the BFA imaging system observes a natural scene characterized by a continuous distribution of spectral radiance. Let $F(x, y, \lambda) \in \mathbb{R}^+$ represent this continuous hyperspectral scene, and $M(x, y, \lambda) \in [0, 1]$ denote the spectral transmission profile of the hardware-implemented broadband filter array.
As light propagates through the optical path, the spectral information is modulated by the mask and subsequently integrated by the monochromatic sensor. The captured 2D intensity measurement $G(x, y)$ is physically formulated as the integral over the responsive spectral range $[\lambda_{min}, \lambda_{max}]$:
\begin{equation}
G(x, y) = \int_{\lambda_{min}}^{\lambda_{max}} F(x, y, \lambda) M(x, y, \lambda) d\lambda + N(x, y)
\label{eq:continuous}
\end{equation}
where $N(x, y)$ accounts for the inherent photon noise and readout noise of the sensor. This equation represents the "ground truth" physics that our system aims to invert.

\subsubsection*{2) Discrete Simulation Framework}
While the ultimate goal is to deploy the algorithm on hardware, training the proposed PF-Trans network requires a massive corpus of paired data (scene-measurement pairs) which is prohibitively expensive to acquire experimentally in large quantities. To address this, we establish a rigorous Forward Physics Simulation framework to generate synthetic training data that statistically aligns with the real-world optical process described above.

We utilize high-quality public hyperspectral datasets to serve as the discrete ground truth $\mathbf{X} \in \mathbb{R}^{H \times W \times C}$, acting as a proxy for the discretized continuous scene. The continuous integral in Eq. \ref{eq:continuous} is numerically approximated by the summation of discrete spectral bands. Consequently, the simulated compressed measurement $\mathbf{Y} \in \mathbb{R}^{H \times W}$ is generated via:
\begin{equation}
\mathbf{Y} = \sum_{k=1}^{C} (\mathbf{X}_k \odot \mathbf{M}_k) + \mathbf{N}
\label{eq:sensing}
\end{equation}
where $\odot$ denotes the element-wise Hadamard product, and $\mathbf{M}_k$ represents the discretized mask pattern at the $k$-th wavelength. 
This simulation strategy effectively prevents "data leakage" by ensuring that the network only sees the compressed $\mathbf{Y}$ (and the known mask $\mathbf{M}$) during inference, while $\mathbf{X}$ is strictly reserved for calculating the supervision loss.
Physics-embedded Input: To guide the reconstruction, we concatenate the simulated measurement $\mathbf{Y}$ with the mask $\mathbf{M}$ to form the input tensor $\mathbf{I}_{in}$. This explicitly informs the network of the spatial-spectral encoding rules applied during the forward simulation, enabling it to mathematically invert the degradation process.

\subsection{Dual-domain Spectral Reconstruction}
\label{sec:network}

Recovering high-fidelity spectra from BFA measurements necessitates resolving two distinct types of signal degradation: the loss of local spectral consistency and the introduction of global periodic artifacts by the mask. Standard Convolutional Neural Networks (CNNs), restricted by their local receptive fields, often fail to differentiate between high-frequency image details and the grid-like aliasing patterns inherent to the BFA. To overcome this dichotomy, we propose the Physics-embedded Frequency-aware Transformer (PF-Trans), which employs a Dual-domain Block (Fig. \ref{fig:architecture}, Middle) to synergize spatial context modeling with frequency-domain de-aliasing.

\subsubsection*{1) Spatial Branch: Local Texture Recovery}
For the spatial pathway, it is imperative to capture long-range dependencies while maintaining computational feasibility. We adopt the Spectral-wise Multi-head Self-Attention (S-MSA) mechanism. Unlike conventional global attention which computes dense maps across spatial pixels, S-MSA operates along the spectral dimension. This design exploits the high correlation between adjacent bands in HSI data, effectively modeling inter-spectral dependencies to recover material signatures. Simultaneously, the residual connections ensure the preservation of high-frequency spatial textures, preventing the over-smoothing often observed in transformer-based architectures.

\subsubsection*{2) Frequency Branch: Global De-aliasing}
Complementary to the spatial branch, the frequency branch is specifically engineered to eliminate mask-induced aliasing. Based on the Convolution Theorem, the periodic grid patterns of the BFA mask manifest as concentrated high-energy spikes in the frequency domain, which are distinct from the natural scene spectrum. Consequently, we separate signal from noise via a global filtering mechanism in the Fourier domain. The input feature $\mathbf{F}_{in}$ is first transformed via 2D Fast Fourier Transform (FFT):
\begin{equation}
\mathbf{Z} = \mathcal{F}(\mathbf{F}_{in}) \in \mathbb{C}^{H \times W \times C'}
\end{equation}
To effectively modulate the complex spectrum, we introduce an Interactive Complex Convolution. Distinct from methods that treat real and imaginary components independently, our approach employs a cross-domain interaction strategy with shared kernels $\mathbf{W}_{r}$ and $\mathbf{W}_{i}$ to strictly emulate complex arithmetic:
\begin{equation}
\left[ \begin{array}{c} \mathfrak{R}(\tilde{\mathbf{Z}}) \\ \mathfrak{I}(\tilde{\mathbf{Z}}) \end{array} \right] = 
\left[ \begin{array}{cc} \mathbf{W}_{r} & -\mathbf{W}_{i} \\ \mathbf{W}_{i} & \mathbf{W}_{r} \end{array} \right] * \left[ \begin{array}{c} \mathfrak{R}(\mathbf{Z}) \\ \mathfrak{I}(\mathbf{Z}) \end{array} \right]
\end{equation}
This operation functions as a global learnable filter that selectively suppresses the frequency components associated with the mask artifacts across the entire image field. The purified features are subsequently restored via Inverse FFT (IFFT):
\begin{equation}
\mathbf{F}_{out} = \mathbf{F}_{in} + \text{LayerNorm}(\mathcal{F}^{-1}(\tilde{\mathbf{Z}}))
\end{equation}
By aggregating features from both branches, the PF-Trans achieves a holistic reconstruction, simultaneously ensuring local spatial fidelity and global spectral cleanliness.

\subsection{Physics-constrained Optimization}
\label{sec:loss}

Deep learning-based inversion methods, while powerful, risk converging to statistically probable but physically invalid local minima, especially when training data is limited. To guarantee the optical plausibility of the results, we impose a closed-loop constraint through a Gray-scale Consistency Loss (Fig. \ref{fig:architecture}, Right).
The optimization objective $\mathcal{L}_{total}$ is composed of a data-fidelity term and a physics-regularization term:
\begin{equation}
\mathcal{L}_{total} = \mathcal{L}_{recon} + \lambda \mathcal{L}_{gray} = \|\hat{\mathbf{X}} - \mathbf{X}_{gt}\|_2^2 + \lambda \mathcal{L}_{gray}
\end{equation}
The term $\mathcal{L}_{gray}$ is derived by re-projecting the reconstructed estimate $\hat{\mathbf{X}}$ through the known forward model:
\begin{equation}
\mathcal{L}_{gray} = \left\| \left( \sum_{k=1}^{C} \hat{\mathbf{X}}_k \odot \mathbf{M}_k \right) - \mathbf{Y} \right\|_2^2
\end{equation}
This loss enforces a rigorous physical constraint: the reconstructed hyperspectral cube, when modulated by the mask, must yield a 2D projection that precisely matches the captured raw measurement $\mathbf{Y}$. This penalizes the network for generating hallucinations that violate the energy conservation laws of the imaging system. We empirically set $\lambda=0.03$ to balance the trade-off between spatial reconstruction quality and physical self-consistency.

% --- 3. EXPERIMENTS ---
\section{Experiments}

\subsection{Experimental Setup}
\textit{1) Datasets and Metrics:}
We evaluate PF-Trans on four remote sensing datasets: GF-5 (HHK and SH scenes), Chikusei, Houston, and a self-collected UAV dataset named KXY. To strictly align with the physical BFA specifications (61 bands, 400-1000 nm) \cite{Bian2024}, we standardized the spectral dimension of all datasets. Specifically, the KXY dataset was acquired using a HeadWall Nano-Hyperspec sensor with 273 raw bands. We performed spectral interpolation and resampling to generate high-fidelity 61-band ground truth, ensuring physical consistency for BFA simulation. The same preprocessing was applied to the satellite and airborne datasets. Performance is evaluated using PSNR$\uparrow$, SSIM$\uparrow$, and SAM$\downarrow$.

\textit{2) Comparative Methods:}
We compare PF-Trans with state-of-the-art methods: the CNN-based \textbf{SR-Net} \cite{Bian2024} and the Transformer-based \textbf{MST++} \cite{Cai2022}. To rigorously validate our contributions, we also conduct a comprehensive ablation study.

% --- TABLE 1: SOTA Comparison ---
\begin{table}[htbp]
\caption{QUANTITATIVE COMPARISON WITH STATE-OF-THE-ART METHODS (MEAN $\pm$ STD).}
\label{tab:comparison}
\centering
\setlength{\tabcolsep}{1.5mm}
\renewcommand{\arraystretch}{1.2}
\begin{tabular}{c|c|ccc}
\toprule
\textbf{Dataset} & \textbf{Method} & \textbf{PSNR}$\uparrow$ & \textbf{SSIM}$\uparrow$ & \textbf{SAM}$\downarrow$ \\
\midrule
% --- GF-5 HHK ---
\multirow{3}{*}{\centering GF-5 HHK} 
 & SR-Net \cite{Bian2024} & 25.05$\pm$0.82 & 0.86$\pm$0.01 & 0.17$\pm$0.02 \\
 & MST++ \cite{Cai2022} & 23.86$\pm$0.31 & 0.90$\pm$0.01 & 0.15$\pm$0.02 \\
 & \textbf{PF-Trans} & \textbf{35.14$\pm$0.35} & \textbf{0.96$\pm$0.01} & \textbf{0.04$\pm$0.01} \\
\midrule
% --- GF-5 SH ---
\multirow{3}{*}{\centering GF-5 SH} 
 & SR-Net \cite{Bian2024} & 33.94$\pm$1.14 & 0.83$\pm$0.03 & 0.16$\pm$0.01 \\
 & MST++ \cite{Cai2022} & 39.56$\pm$2.17 & 0.96$\pm$0.02 & 0.17$\pm$0.02 \\
 & \textbf{PF-Trans} & \textbf{48.50$\pm$4.27} & \textbf{0.98$\pm$0.03} & \textbf{0.07$\pm$0.01} \\
\midrule
% --- Chikusei ---
\multirow{3}{*}{\centering Chikusei} 
 & SR-Net \cite{Bian2024} & 33.56$\pm$2.19 & 0.85$\pm$0.05 & 0.16$\pm$0.02 \\
 & MST++ \cite{Cai2022} & 37.73$\pm$2.08 & 0.93$\pm$0.03 & 0.10$\pm$0.04 \\
 & \textbf{PF-Trans} & \textbf{41.91$\pm$1.79} & \textbf{0.97$\pm$0.02} & \textbf{0.06$\pm$0.02} \\
\midrule
% --- Houston ---
\multirow{3}{*}{\centering Houston} 
 & SR-Net \cite{Bian2024} & 31.02$\pm$0.09 & 0.81$\pm$0.01 & 0.17$\pm$0.02 \\
 & MST++ \cite{Cai2022} & 35.80$\pm$0.07 & 0.90$\pm$0.02 & 0.18$\pm$0.01 \\
 & \textbf{PF-Trans} & \textbf{38.16$\pm$0.09} & \textbf{0.94$\pm$0.01} & \textbf{0.11$\pm$0.01} \\
\midrule
% --- KXY ---
\multirow{3}{*}{\centering KXY} 
 & SR-Net \cite{Bian2024} & 27.63$\pm$2.91 & 0.81$\pm$0.10 & 0.18$\pm$0.03 \\
 & MST++ \cite{Cai2022} & 36.96$\pm$2.71 & 0.95$\pm$0.02 & 0.09$\pm$0.02 \\
 & \textbf{PF-Trans} & \textbf{42.07$\pm$2.53} & \textbf{0.97$\pm$0.02} & \textbf{0.05$\pm$0.01} \\
\bottomrule
\end{tabular}
\end{table}

\subsection{Performance Analysis}
\textit{1) Comparison with SOTA:}
Table \ref{tab:comparison} reports the quantitative results. PF-Trans consistently outperforms existing methods across all datasets. Notably, on the urban GF-5 SH scene, our method achieves a remarkable PSNR of 48.50 dB, surpassing the MST++ baseline by nearly 9 dB. This significant boost indicates that our frequency-aware design effectively handles the complex high-frequency details typical of urban environments, which are often confused with mask aliasing in spatial-only models.

% --- TABLE 2 & TABLE 3: Continuous Layout ---
\begin{table}[htbp]
\caption{ABLATION SETTINGS: COMPONENT CONFIGURATIONS.}
\label{tab:ablation_setup}
\centering
\setlength{\tabcolsep}{3.5mm}
\renewcommand{\arraystretch}{1.2}
\begin{tabular}{l|ccc}
\toprule
Model Variant & $\mathcal{L}_{gray}$ & Mask Injection & Freq. Branch \\
\midrule
Baseline (MST++) & & & \\
+ Gray Loss & \checkmark & & \\
+ Mask Injection & \checkmark & \checkmark & \\
\textbf{PF-Trans} & \checkmark & \checkmark & \checkmark \\
\bottomrule
\end{tabular}
\end{table}

\begin{table}[htbp]
\caption{ABLATION RESULTS AVERAGED OVER 5 DATASETS.}
\label{tab:ablation_results}
\centering
\setlength{\tabcolsep}{4.0mm}
\renewcommand{\arraystretch}{1.2}
\begin{tabular}{l|ccc}
\toprule
Method & PSNR$\uparrow$ & SSIM$\uparrow$ & SAM$\downarrow$ \\
\midrule
Baseline & 34.78 & 0.93 & 0.14 \\
+ Gray Loss & 36.10 & 0.94 & 0.12 \\
+ Mask Injection & 38.37 & 0.94 & 0.10 \\
\textbf{PF-Trans} & \textbf{41.16} & \textbf{0.96} & \textbf{0.07} \\
\bottomrule
\end{tabular}
\end{table}

\textit{2) Ablation Study:}
To validate our contributions, we conduct a progressive ablation study averaged across all five datasets. The configurations and results are presented in Table \ref{tab:ablation_setup} and Table \ref{tab:ablation_results}, respectively.

\textbf{Effect of Physics Embedding:} Compared to the Baseline (MST++), the introduction of $\mathcal{L}_{gray}$ (+ Gray Loss) improves the average PSNR from 34.78 dB to 36.10 dB. Further injecting the physical mask pattern (+ Mask Injection) yields a substantial boost to 38.37 dB. This confirms that explicitly embedding the physical sensing model effectively narrows the solution space.

\textbf{Effect of Frequency Branch:} Finally, the full PF-Trans (integrating the frequency branch) achieves the best performance with an average PSNR of 41.16 dB. This demonstrates that the Dual-domain Block is critical for decoupling stubborn periodic artifacts that spatial-only models fail to remove.

% --- FIGURE 2: Frequency Comparison (on KXY) ---
\begin{figure}[htbp]
\centering
\includegraphics[width=\columnwidth]{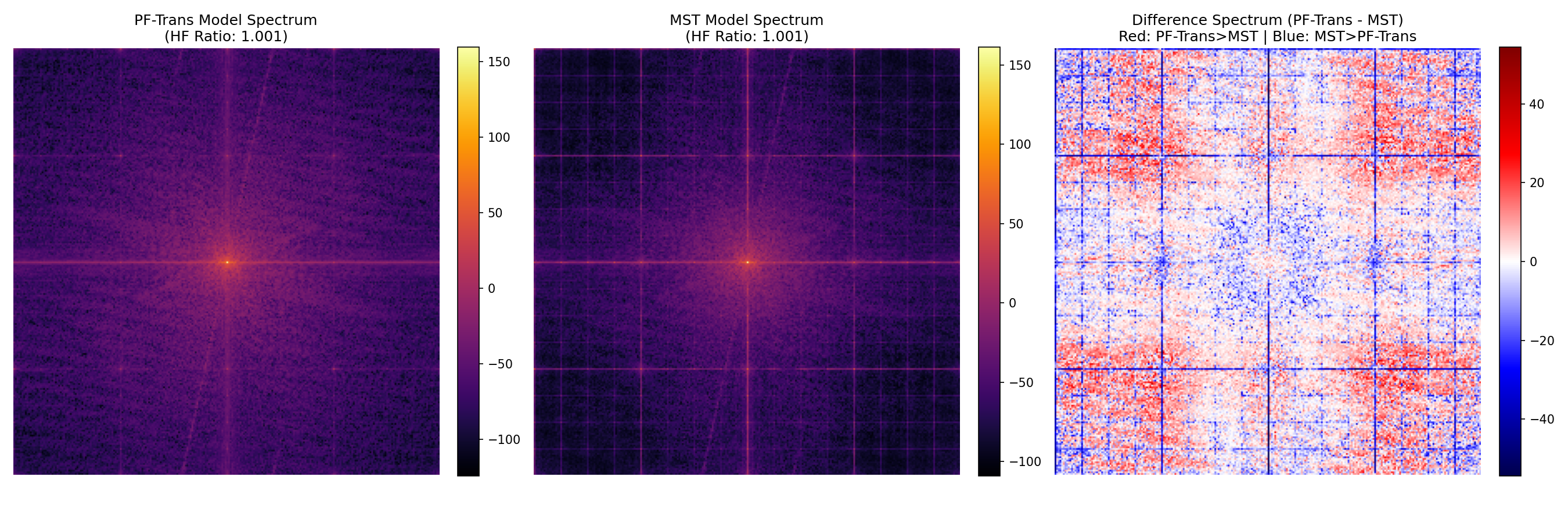}
\caption{Visual quality comparison on the KXY dataset. Top row: Reconstructed spectral images and corresponding error maps. Bottom row: Log-amplitude frequency spectra. MST++ exhibits residual grid-like artifacts and high-frequency spectral spikes. In contrast, PF-Trans effectively suppresses these aliases via the frequency branch, yielding higher physical fidelity.}
\label{fig:freq_comparison}
\end{figure}

% --- FIGURE 3: Spectral Fidelity Comparison (on GF-5 HHK) ---
\begin{figure}[htbp]
\centering
% Updated dataset name in caption to GF-5 HHK
\includegraphics[width=\columnwidth]{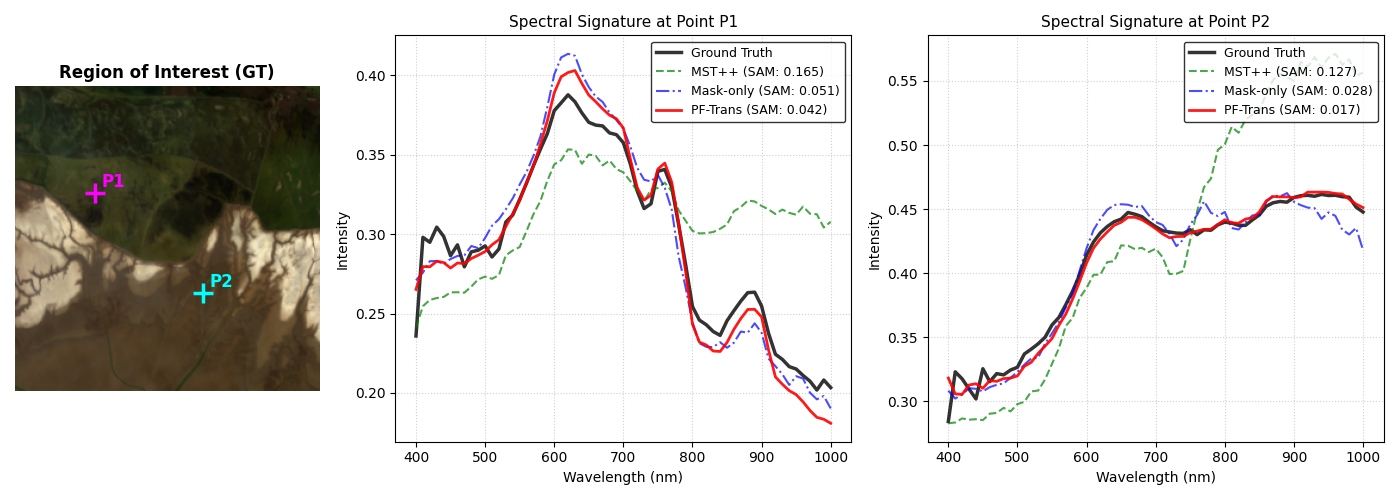}
\caption{Spectral fidelity comparison on the GF-5 HHK dataset. Two representative points (P1 and P2) are selected from the ROI (left). The spectral curves (right) demonstrate that our PF-Trans (red line) achieves the highest alignment with the Ground Truth (black line), yielding the lowest Spectral Angle Mapper (SAM) scores compared to MST++ and the Mask-only variant.}
\label{fig:spectral_curves}
\end{figure}

\textit{3) Visual and Spectral Analysis:}
Fig. \ref{fig:freq_comparison} visualizes the efficacy of the frequency branch on the KXY dataset. The frequency spectrum of MST++ contains anomalous high-frequency energy spikes caused by mask aliasing. In contrast, PF-Trans effectively suppresses these specific spectral spikes, resulting in a cleaner spectrum and a spatially smooth error map, confirming the necessity of frequency-domain processing.

To further verify the spectral fidelity, Fig. \ref{fig:spectral_curves} compares the reconstructed spectral signatures of two representative spatial points from the GF-5 HHK dataset. As observed, the baseline MST++ suffers from significant spectral distortions, with SAM values of 0.165 and 0.127 for points P1 and P2, respectively. The introduction of physics embedding (Mask-only) drastically reduces these errors. Furthermore, our full PF-Trans model achieves the best match with the ground truth curves, reducing the SAM values to as low as 0.042 and 0.017. This near-perfect alignment confirms that the proposed frequency-aware architecture not only removes spatial artifacts but also preserves the precise physical spectral properties of the scene.

% --- 4. CONCLUSION ---
\section{Conclusion}
In this paper, we proposed PF-Trans for BFA spectral reconstruction, explicitly addressing mask-induced aliasing through a physics-embedded framework. By integrating mask injection with a frequency-aware Dual-domain Block, our method effectively decouples global periodic artifacts from local ground textures. Extensive evaluations across satellite, airborne, and UAV platforms demonstrate that PF-Trans achieves state-of-the-art performance, notably reaching 48.50 dB on the GF-5 Shanghai dataset. Future work will focus on exploring lightweight optimizations to facilitate real-time onboard processing for edge computing applications.

\clearpage
\small
\bibliographystyle{IEEEtranN}
\bibliography{sample}

\end{document}